\documentclass[conference]{IEEEtran}

\DeclareUnicodeCharacter{211D}{\,}
\usepackage[T1]{fontenc}    
\usepackage{hyperref}       
\usepackage{url}            
\usepackage{booktabs}       
\usepackage{amsfonts}       
\usepackage{nicefrac}       
\usepackage{microtype}      
\usepackage{lipsum}

\usepackage{filecontents}
\usepackage{comment}
\usepackage{enumerate}
\usepackage{amsfonts}
\usepackage{amssymb}
\usepackage{verbatim}
\usepackage{setspace}
\usepackage{mathrsfs}
\usepackage{tikz}
\usetikzlibrary{arrows,shapes}
\usepackage{graphicx}

\usepackage{subcaption}

\usepackage{etoolbox}

\usepackage{todonotes}
\usepackage{bm} 
\usepackage{float}
\usepackage{graphicx}
\usepackage{tikz}
\usetikzlibrary{shapes.geometric}
\usetikzlibrary{arrows}
\usetikzlibrary{arrows.meta,calc,decorations.markings,math,arrows.meta}

\usepackage{amsmath}
\usepackage{mathdots}
\usepackage{diagbox}
\usepackage{amsfonts}
\usepackage{amssymb}

\begin{document}
%
\title{
BundledSLAM: An Accurate Visual SLAM System Using Multiple Cameras}

\author{\IEEEauthorblockN{Han Song}
\IEEEauthorblockA{
University of Southern California\\
Los Angeles, California, 90007\\
Email: hsong427@usc.edu}
\and
\IEEEauthorblockN{Cong Liu}
\IEEEauthorblockA{Peng Cheng laboratory\\
Shenzhen, China, 518066\\
Email: liuc@pcl.ac.cn}
\and
\IEEEauthorblockN{Huafeng Dai}
\IEEEauthorblockA{Cornell University\\
Ithaca, New York, 14850 \\
Email: hd338@cornell.edu}
}


%


\maketitle

\begin{abstract}
Multi-camera SLAM systems offer a plethora of advantages, primarily stemming from their capacity to amalgamate information from a broader field of view, thereby resulting in heightened robustness and improved localization accuracy. In this research, we present a significant extension and refinement of the state-of-the-art stereo SLAM system, known as ORB-SLAM2, with the objective of attaining even higher precision. To accomplish this objective, we commence by mapping measurements from all cameras onto a virtual camera termed BundledFrame. This virtual camera is meticulously engineered to seamlessly adapt to multi-camera configurations, facilitating the effective fusion of data captured from multiple cameras. Additionally, we harness extrinsic parameters in the bundle adjustment (BA) process to achieve precise trajectory estimation.Furthermore, we conduct an extensive analysis of the role of bundle adjustment (BA) in the context of multi-camera scenarios, delving into its impact on tracking, local mapping, and global optimization. Our experimental evaluation entails comprehensive comparisons between ground truth data and the state-of-the-art SLAM system. To rigorously assess the system's performance, we utilize the EuRoC datasets. The consistent results of our evaluations demonstrate the superior accuracy of our system in comparison to existing approaches.
\end{abstract}


\IEEEpeerreviewmaketitle

\section{Introduction}
\label{sec:intro}
Compared to the extensive research on monocular SLAM systems, there are relatively few Visual-Inertial Odometry (VIO) solutions designed for multi-camera SLAM systems.In many robot applications and for Micro Aerial Vehicles (MAVs), having a wide field of view (FoV) is crucial for effective perception capabilities. However, the predominant focus in current research on Visual Simultaneous Localization and Mapping (SLAM) primarily centers around monocular, stereo, and RGBD cameras. These visual SLAM systems may face challenges such as limited FoV and a single orientation, which can negatively impact their robustness and accuracy due to limited visual data collection.A significant advantage of multi-camera SLAM systems lies in their wide FoV. This characteristic not only addresses the robustness and accuracy issues found in previous SLAM systems but also enhances the efficiency of map construction. For instance, in a multi-camera SLAM system, if certain cameras are obstructed or malfunctioning, the remaining cameras can continue to function normally and provide an ample supply of 3D data points for map generation. 
\begin{figure}[ht]  
\centering
\includegraphics[width=1.0\columnwidth]{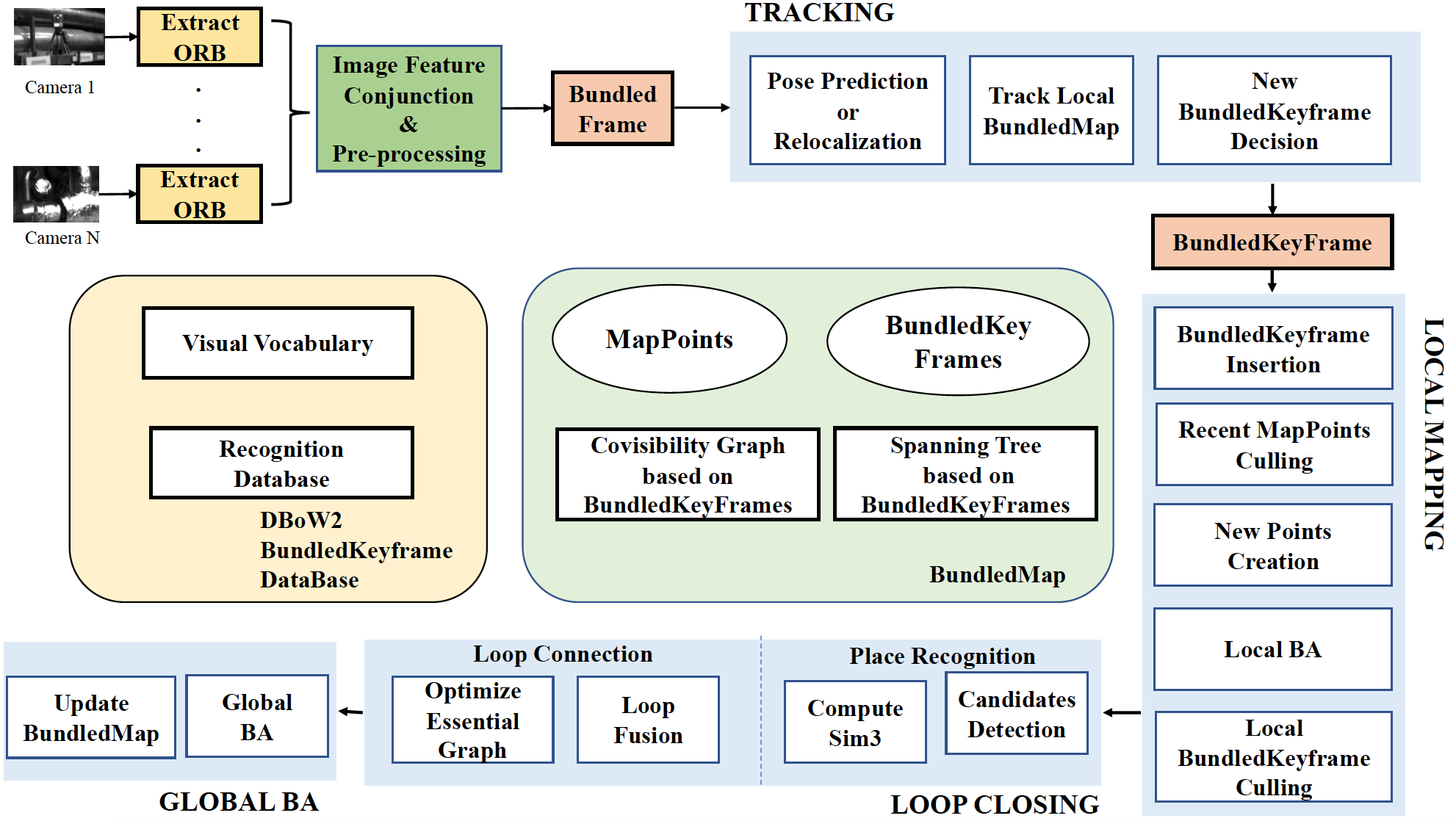}
\caption{Pipeline of BundledSLAM}
\label{fig:pipeline}
\end{figure}
Moreover, the amalgamation of image information from multiple cameras expands the opportunities for feature matching and place recognition, leading to the identification of the most reliable image features. Early research in multi-camera SLAM primarily concentrated on the field of Structure from Motion (SFM). Pless~\cite{1211520} first delve into the theoretical utilization of multiple camera systems in SFM. He deduce a generalized camera model and epipolar constraints, outlining how to implement multi-camera-based offline SFM for self-driving applications. This marked the inaugural use of a multi-camera system in the SFM.
Building on this, Frahm et al.~\cite{frahm2004pose} introduce the concept of representing multiple cameras with a virtual camera. They achieve pose estimation for multiple camera systems using this approach, even for systems with non-overlapping fields of view. Sola et al.~\cite{4631502} explore the adaptation of single-camera SLAM algorithms for multi-camera SLAM and develope methods for the online automatic calibration of multi-view SLAM systems.

Harmat~\cite{7} improve the perspective camera model in PTAM by introducing a generalized polynomial model. He equippe a small drone platform with a pair of ground-facing stereo cameras and another wide-angle camera facing in the opposite direction to assess the performance enhancement of PTAM in indoor environments using multiple cameras. Subsequently, Harmat develope the classical SLAM system MCPTAM \cite{8} based on his improved PTAM work. This SLAM system achieve algorithm convergence without requiring precise scale information, thanks to the use of an omnidirectional vision Taylor model and a spherical coordinate update method. The concept of multiple key frames (MKF) was introduced in this work.

Tribou \cite{9} also worke on methods involving non-overlapping fields of view to enhance the robustness of drone pose estimation based on PTAM. Harmat's research extended into multi-camera pose estimation. Yang et al. \cite{10} employe a common perspective camera model to achieve robust pose tracking on autonomous micro aerial vehicle (MAV) platforms. They demonstrate the flexibility of configuring the number and orientation of cameras and integrating visual information from various perspectives.

Furthermore, both \cite{trvomcs} and \cite{svo} extend their approaches for multi-camera systems. They employe two parallel threads for estimating and mapping while minimizing photometric errors.
In this paper, our primary objective is to enhance accuracy by extending the capabilities of ORB-SLAM2 as described in \cite{ORBSLAM2} to incorporate pose estimation and map reuse from multiple cameras. All image features from these multiple cameras will be amalgamated to create comprehensive data for feature matching in the tracking module and place recognition during loop closure. Furthermore, we achieve pose updates and optimization by minimizing a cost function involving multiple cameras.

Taking inspiration from work of Wang et al. in treating visual SLAM systems as a virtual sensor~\cite{EDI}. We initiate the process by mapping measurements from all cameras onto a virtual camera called BundledFrame. This virtual camera is designed to seamlessly accommodate multi-camera setups, allowing us to efficiently combine data from multiple cameras and then apply bundle adjustment with extrinsic parameters to optimize pose in a multi-camera SLAM system.
The major contributions of our multi-camera system can be summarized by combining the following key features:
\begin{itemize}
\item Comprehensiveness: We provide a complete SLAM system for multiple cameras, encompassing loop closure and map reuse.
\item Extensibility: By leveraging an efficient data structure known as "Bundled," we consolidate data from multiple cameras into a "BundledFrame" or "BundledKeyframe." This forms the bedrock for all system operations, encompassing tracking, place recognition, and optimization. Our system readily adapts to additional cameras by implementing Bundle Adjustment (BA) with extrinsic parameters between the cameras.
\end{itemize}

\section{BundledSLAM}
\label{sec:BundledSLAM}
\subsection{System Overview}\label{sec: System}
\begin{figure}[htp]  
\centering
\includegraphics[width=1\linewidth, height=2in]{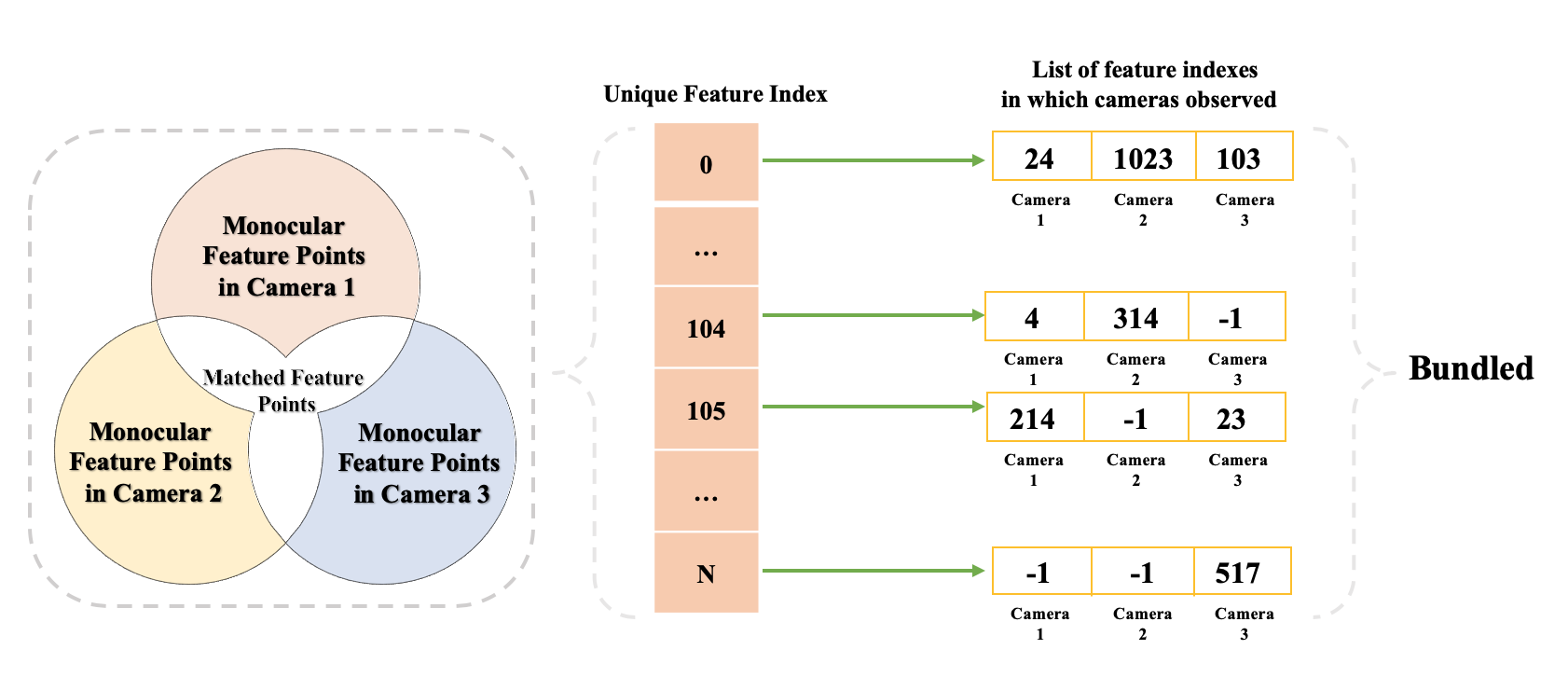}
\caption{An example of Bundled, comprising unique feature IDs, monocular points, and matched points.\\
Number of Unique Feature = Number of Monocular Feature + Number of Matched Feature}
\label{fig:example_feature}
\end{figure}

Figure~\ref{fig:pipeline} illustrates the pipeline of our multi-camera SLAM system. Our system is organized into three primary parallel threads: tracking, local mapping, and loop closing.
The tracking module is responsible for estimating incremental motion by identifying feature matches in the local BundledMap and minimizing re-projection errors using our motion-only Bundle Adjustment (BA). It also determines if the current frame qualifies as a new BundledKeyframe, which is subsequently integrated into the local mapping thread.
The local mapping thread manages new BundledKeyframes, involving consistent connection updates, new map point creation, and the removal of redundant data. It optimizes the local BundledMap through the implementation of our local Bundle Adjustment (BA).
The primary goal of the loop closing thread is to detect significant loops and perform pose-graph optimization. Additionally, this thread initiates another thread to conduct a global Bundle Adjustment (BA) aimed at correcting accumulated errors.
BundledSLAM, like previous feature-based SLAM systems, begins by preprocessing the input images to extract features at salient keypoint locations. Once this feature extraction is complete, the original input images are discarded, and all subsequent system operations rely on these features for tasks such as tracking, place recognition, and optimization.
To facilitate the integration of image features from various cameras, we implement feature matching across multiple cameras to assign a unique feature ID to each feature point. The associations between these unique feature IDs and their corresponding camera IDs (each observed camera) are stored within our dedicated data structure named 'Bundled,' as depicted in Figure~\ref{fig:example_feature}. Feature points observed exclusively by a single camera are referred to as 'monocular feature points,' while the others are termed 'matched points.'
Bundled provides a wide range for search of feature than ORB-SLAM2, which only considers matching between the left-to-left camera, to find out a accurate image feature and map points.
\subsection{BundledFrame, BundledKeyframe and BundledMap}
\begin{figure}[H]  
\centering
\includegraphics[width=1.0\columnwidth]{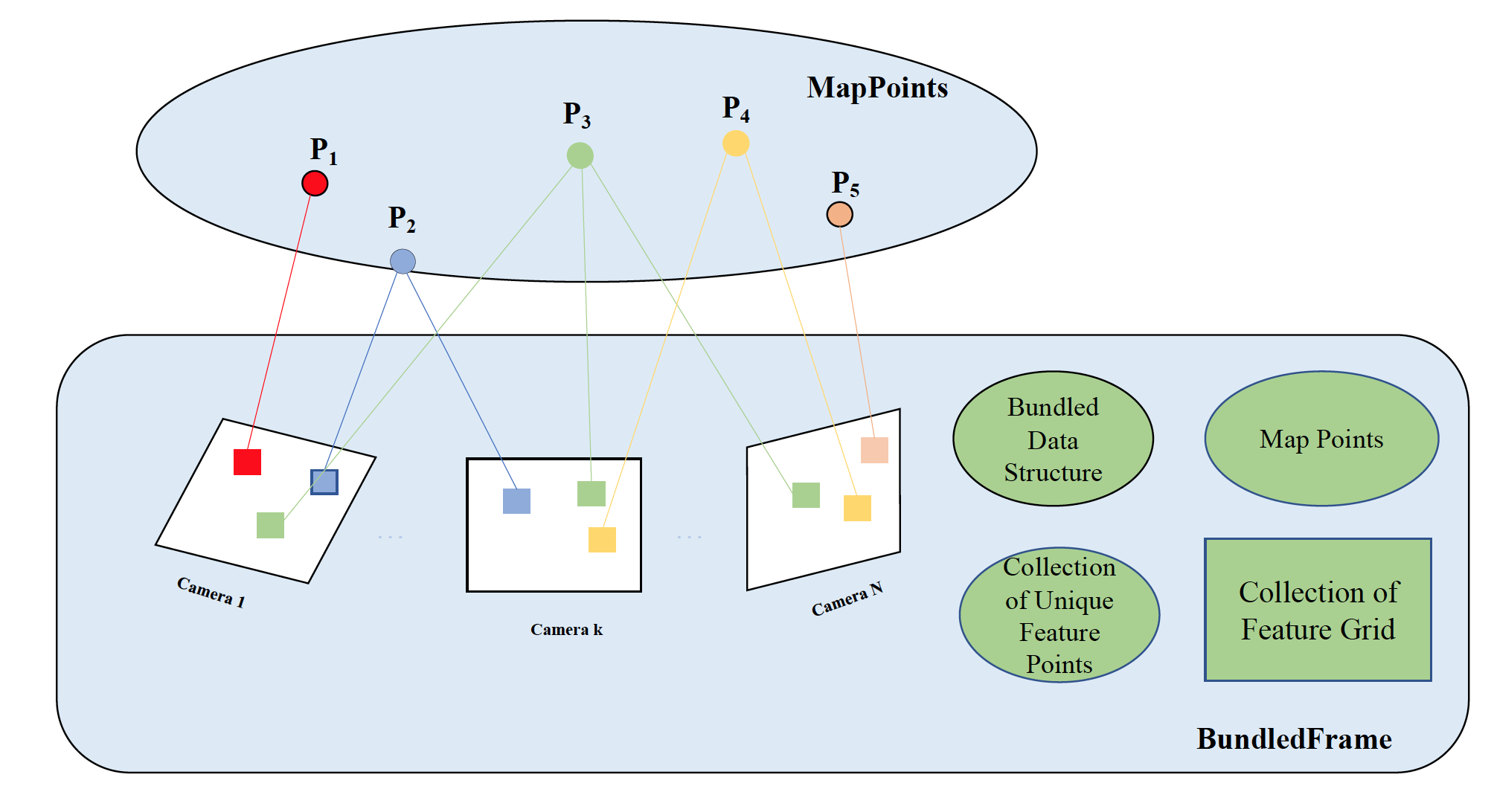}
\caption{General Overview of BundledFrame}
\label{fig:bundledframe}
\end{figure}
\textbf{BundledFrame} (See  Figure~\ref{fig:bundledframe}) includes all image information from different cameras. In addition, frames from different cameras are divided into $64\times48$ size grid separately, and features are allocated to corresponding grids according to positions in order to reduce the time complexity of matching.\par
\textbf{BundledKeyframe} is decided in the last step of tracking if the current BundledFrame is selected as new BundledKeyframe. Meanwhile, BundledKeyframe is the basic operation unit in local mapping and local closing. Covisibility is represented as an undirected weighted graph as in \cite{covisib}. An Co-visible edge between two Bundledkeyframes exists if they share observations of the same map points. The following Local BA and loop closing depend on these covisibility information.\par
\textbf{BundledMap} consists of all BundledKeyframes and a collection of all map points that have been seen by BundledKeyframes.
\subsection{Camera Projection Model with Multiple Cameras}
The $i^{th}$ camera state vector at time-step k, $\bm{c}_i^k$, consists of the sensor orientation and position with respect to a world coordinates:
\begin{equation}
\bm{c}_i^k = \left [
 \begin{matrix}
   \bm{R}_{iw}^k & \bm{t}_{iw}^k \\
   \bm{0} & \bm{1} 
  \end{matrix}
  \right  ]
\end{equation}
where $\bm{R}_{iw}^k \in SO(3)$ is the rotation matrix from the world coordinates to the $i^{th}$ camera coordinates and position $\bm{t}_{iw}^k \in \mathbb{R}^3$.
Using the same calibrated camera projection model as in \cite{ORBSLAM2}, the observation of feature $j$ is described by the image project of the 3D feature position vector in world coordinates $\bm{P}_j \in \mathbb{R}^3$ to the $i^{th}$ camera:
\begin{equation}
\bm{z}_{ji} = \bm{h}(\bm{R}_{iw}^k\bm{P}_j+\bm{t}_{iw}^k) + \bm{n}_{ji}
\end{equation}
where $\bm{h}(\cdot)$ is the $i^{th}$ camera projection function, and \bm{$n_{ji}} \sim \mathcal{N}(\bm{0}_{2 \times 1}, \bm{Q}_{ji})$ is the measurement noise vector, modeled as zero-mean Gaussian variable with covariance matrix \bm{$Q}_{ji}$.\par
Since the primary objective of the SLAM system is to estimate the state, we calculate the pose update for a specified camera at time-step $k$, which we refer to as the first camera $C_1^k$. This update is based on measurements from all cameras at the same time. The poses of other cameras can then be updated through an estimated constant transformation relative to $C_1^k$. 
\begin{equation}
    \bm{T}_{i1} = \left [
 \begin{matrix}
   \bm{R}_{i1}^k & \bm{t}_{i1}^k \\
   \bm{0} & \bm{1} 
  \end{matrix}
  \right  ]
\end{equation}

\begin{equation}
    \bm{c}_i^k = \bm{T}_{i1}\bm{c}_1^k
\end{equation}
where $\bm{R}_{i0} \in SO(3)$ is the rotation matrix from the first frame to the $i^{th}$ camera frame and translation $\bm{t}_{i0} \in \mathbb{R}^3$, and the constant relative transformation $\bm{T}_{i1} \in SE(3)$. \par
The reason for setting $\bm{T}_{i1}$ as a constant is that it is unobservable. We fix relative transformation $\bm{T}_{i0}$ in order to avoid the drift of our whole system caused by the variables($\bm{c}_i^k$,  $\bm{T}_{i1}$) moving freely in zero spaces. 
\subsection{Motion Estimation and Bundle Adjustment with Multiple Cameras}
Our motion estimation approach, as proposed, aims to estimate the variable $\bm{c}_1^k$ at time step $k$. Initially, $\bm{c}_1^k$ is assigned an initial value. If the tracking was successful in the previous frame, we set the initial value equal to the previous relative motion, assuming a constant velocity motion model. Subsequently, pose update and optimization are performed. The camera pose update and bundle adjustment are divided into three parts: Motion-only Bundle Adjustment (BA), Local BA, and Global BA.

In our system with N cameras, we redefine the cost function to accommodate multiple cameras and minimize it. The minimization of the nonlinear cost function can be achieved through iterative methods such as Gauss-Newton \cite{GN} or Levenberg-Marquardt methods in g2o \cite{g2o}.
\par
\begin{figure}[H]  
\centering
\includegraphics[width=1.0\columnwidth]{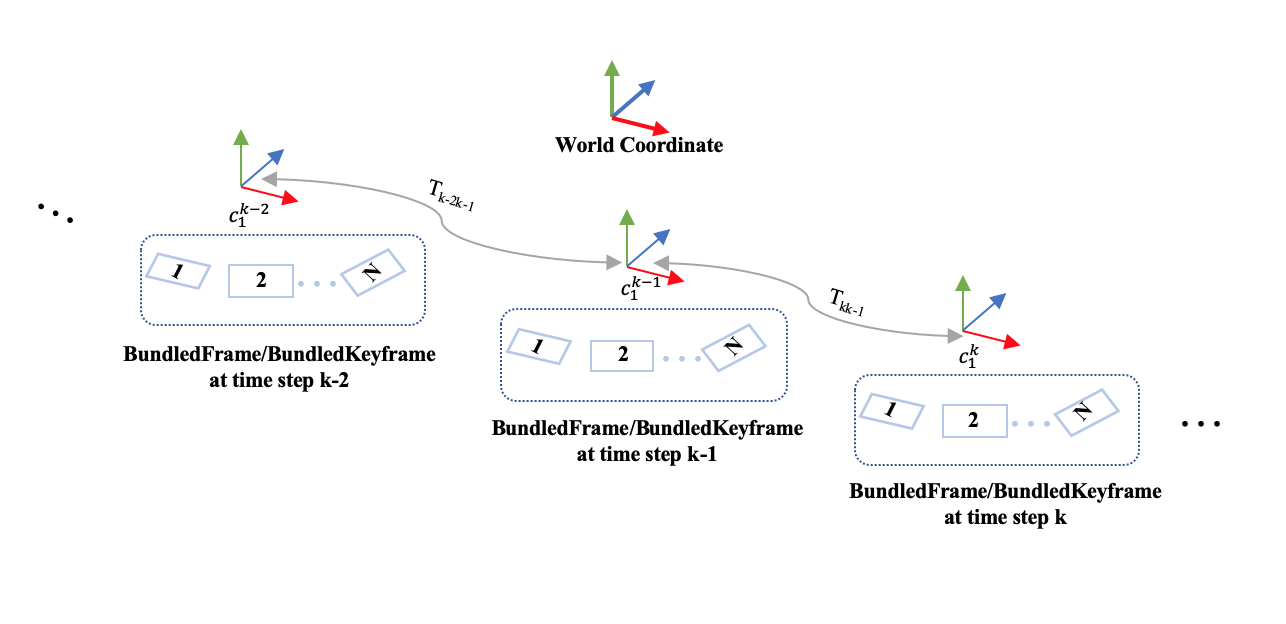}
\caption{Visual odometry with multiple synchronized cameras. The goal is to estimate the relative motion of the first camera $\bm {C}_1^k$ in BundledFrame or BundledKeyframe at each moment to world coordinates.}
\label{fig:graph}
\end{figure}
\textbf{Motion-only BA} performs to find optimal camera $C_1^k$ pose $\bm{c}_1^k$ in the tracking thread. In our $N$-cameras system, we define our cost function(see equation~\eqref{eq:cost}) and minimize the reprojection error between matched 3D map points $\bm{P}_j$ in world coordinates and its corresponding feature points $\bm{u}_{ji}^k$ observed by $i^{th}$ camera at time-step $k$: 
\begin{equation}\label{eq:cost}
    \bm{e_{ji}^k} = \bm{u}_{ji}^k - \bm{h}(\bm{R}_{i1}^k(\bm{R}_{1w}^k\bm{P}_j+\bm{t}_{1w}^k) + \bm{t}_{i1}^k)
\end{equation}

\begin{equation}
\bm{c_1^k} = \{\bm{R}_{1w}^k, \bm{t}_{1w}^k\} = \mathop{\arg\min}_{\bm{R}_{1w}^k, \bm{t}_{1w}^k}  \sum_{j \in S_i^k}\sum_{i \in N} \rho(\left \| \bm{e_{ji}^k} \right \|_{\bm{Q}_{ji}}^2)
\end{equation}

where $\rho$ is the robust Huber cost function and $S_i^k$ is set of all matches of $i^{th}$ camera at time-step $k$. \par
We use the monocular projection functions $h(\cdot)$ for our $N$-cameras instead of rectified stereo projection for dual cameras in ORB-SLAM2. The monocular projection are defined as follows:
\begin{equation}
    h \left (\left [
 \begin{matrix}
  X_i^k \\
  Y_i^k\\
  Z_i^k
  \end{matrix}
  \right  ]\right) = \left [
 \begin{matrix}
  f_x\frac{X_i^k}{Z_i^k} + c_x\\
  \\
  f_y\frac{Y_i^k}{Z_i^k} + c_y
  \end{matrix}
  \right  ]
\end{equation}
where $f_x$ and $f_y$ are the focal length and $c_x$ and $c_y$ are the principal point. Vector $[X_i^k,Y_i^k,Z_i^k]^\mathsf{T}$ is a feature position in $i^{th}$ camera coordinates at time-step $k$.
For the pose update of our $N$-cameras system, the optimization problem is to find out the optimal camera $C_1^k$ pose update $\Delta\bm{c}_1^k$ which is a small disturbing quantity at time-step $k$. We compute the jacobian matrix of $\bm{e_{ji}^k}$ with respect to the estimated $C_1^k$ pose update $\Delta\bm{c}_1^k$:
\begin{equation}
    \frac{\partial \bm{e_{ji}^k}}{\partial\bm{c}_1^k} = -\left [
 \begin{matrix}
  \frac{f_x}{Z_i^k} & 0 & -\frac{f_xX_i^k}{Z_i^2} \\
  0 & \frac{f_y}{Z_i^k} & -\frac{f_yY_i^k}{Z_i^2}
  \end{matrix}
  \right  ]\bm{R_{i1}^k}\left [ \left \lfloor -\bm{P}_j^\prime \right \rfloor_\times \quad \bm{I}_3\right]
\end{equation}
where $\bm{P}_j^\prime \in \mathbb{R}^3$ is a feature ($j$) position vector in camera $C_1^k$ coordinates, $\bm{I_3}$ denotes the $3 \times 3$ identity matrix, and $\left \lfloor \bm{P}_j^\prime \right \rfloor_\times$ is the skew-symmetric matrix associated with the vector $\bm{P}_j^\prime$.

\textbf{Local BA} is responsible for optimizing a set of co-visible bundled keyframes denoted as $\mathcal{B_L}$ and all the points observed in these bundled keyframes, represented by $\mathcal{P_L}$. To ensure that variables do not converge to a zero space, we employ the same strategy as ORB-SLAM2. All other bundled keyframes $\mathcal{B_F}$ that do not belong to $\mathcal{B_L}$ but observe points in $\mathcal{P_L}$, contribute to the cost function while remaining fixed during optimization \cite{ORBSLAM2}. Additionally, we calculate the Jacobian matrix of $\bm{e_{ji}^k}$ with respect to the estimated pose update $\Delta\bm{c}_1^k$ (as seen in equation~\eqref{eq:jocobian_mp}) and the map point $\bm{P}_j$:
\begin{equation}\small\label{eq:lba}
\begin{aligned}
    \{\bm{P}_j, \bm{R}_{lw}^k, \bm{t}_{lw}^k | j &\in \mathcal{P_L}, l \in \mathcal{B_L} \} \\
&=  \mathop{\arg\min}_{\bm{P}_j, \bm{R}_{1w}^k, \bm{t}_{1w}^k} \sum_{b \in \mathcal{B_L} \cup \mathcal{B_F} }  \sum_{j \in S_i^k}\sum_{i\in N}\rho(\left \| \bm{e_{ji}^k} \right \|_{\bm{Q}_{ji}}^2)
\end{aligned}
\end{equation}

\begin{equation}\label{eq:jocobian_mp}
    \frac{\partial \bm{e_{ji}^k}}{\partial\bm{P}_j} = -\left [
 \begin{matrix}
  \frac{f_x}{Z_i^k} & 0 & -\frac{f_xX_i^k}{Z_i^2} \\
  0 & \frac{f_y}{Z_i^k} & -\frac{f_yY_i^k}{Z_i^2}
  \end{matrix}
  \right  ]\bm{R_{i1}^k} \bm{R}_{1w}^k
\end{equation}
\par
\textbf{Global BA} optimizes all bundledkeyframes and map points in the Bundledmap, except the origin bundledkeyframes by using the same optimization strategy as Local BA.
\subsection{Loop Closing}
Loop closing encompasses three key steps: loop detection, similarity transformation computation, and loop fusion. In BundledSLAM, loop detection relies on querying the database, employing a bag-of-words place recognition module based on DBoW2 \cite{DBoW}. We create a visual vocabulary using ORB descriptors extracted from a large dataset of images \cite{vocab} to ensure robust performance across different environments with the same vocabulary.
Every unique feature descriptor in our system is assigned to a specific visual word within the vocabulary. Unlike ORB-SLAM2, BundledSLAM incrementally builds a database based on BundledKeyframe information, including an inverted index. This index keeps track of which BundledKeyframes have observed each visual word in the vocabulary. Consequently, our BundledKeyframe database offers a broader search scope for loop candidates, resulting in more accurate loop closure than what ORB-SLAM2 achieves.
Furthermore, when querying the recognition database, the similarity between the bag-of-words vector of the current BundledKeyframe $B_i$ and all its neighbors in the covisibility graph is computed, and we set a threshold score $s_{min}$. Our BundledSLAM is more stringent in loop candidate detection than ORB-SLAM2, as it discards all BundledKeyframes with a score lower than $s_{min}$. This rigorous approach enhances the precision of loop closure in our system.
\begin{figure*}[ht]
\centering
\begin{minipage}{0.7\textwidth}
\centering
    \includegraphics[width=0.9\linewidth, trim=2.2cm 2.4cm 2cm 0cm]{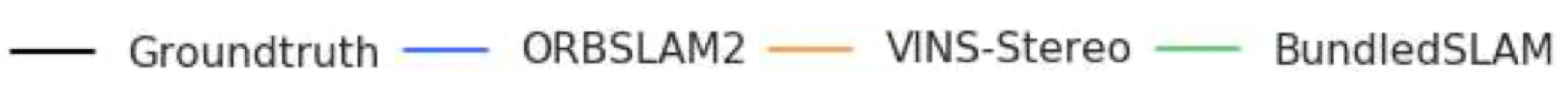}
\end{minipage}\\
\begin{minipage}{1\textwidth}
\begin{subfigure}{.19\textwidth}
  \centering
   \vspace{1cm}
  \includegraphics[width=0.98\linewidth, trim=1cm 0cm 0cm 1cm]{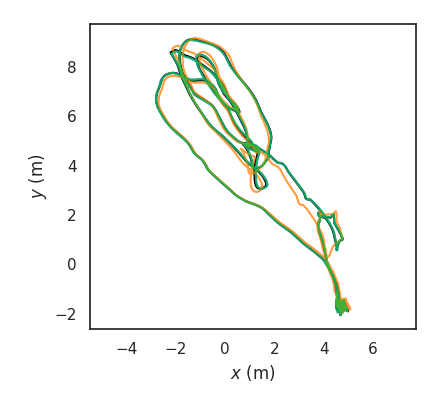} 
  \caption{MH01 Sequence}
\end{subfigure}
\begin{subfigure}{.19\textwidth}
  \centering
     \vspace{1cm}
  \includegraphics[width=0.98\linewidth, trim=1cm 0cm 0cm 1cm]{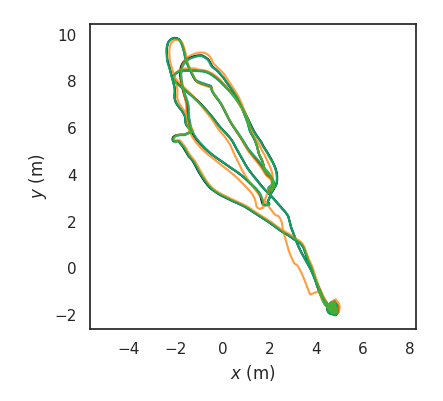} 
  \caption{MH02 Sequence}
\end{subfigure}
\begin{subfigure}{.19\textwidth}
  \centering
     \vspace{1cm}
  \includegraphics[width=0.98\linewidth, trim=1cm 0cm 0cm 1cm]{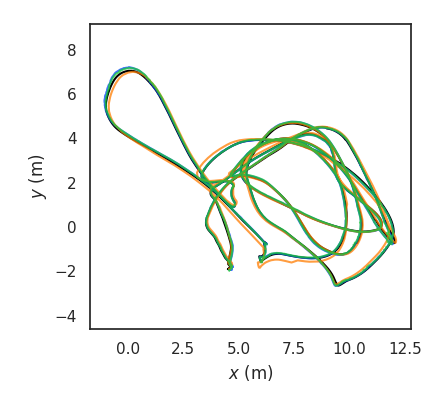} 
  \caption{MH3 Sequence}
\end{subfigure}
\begin{subfigure}{.19\textwidth}
  \centering
     \vspace{1cm}
  \includegraphics[width=0.98\linewidth, trim=1cm 0cm 0cm 1cm]{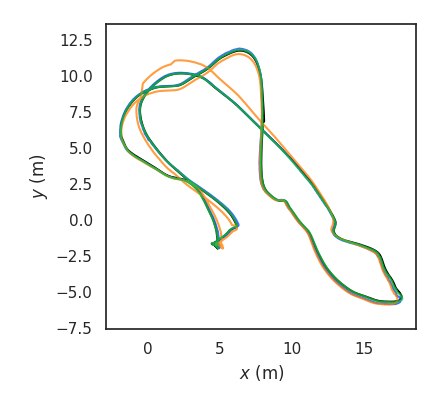} 
  \caption{MH04 Sequence}
\end{subfigure}
\begin{subfigure}{.19\textwidth}
  \centering
    \vspace{1cm}
  \includegraphics[width=0.98\linewidth, trim=1cm 0cm 0cm 1cm]{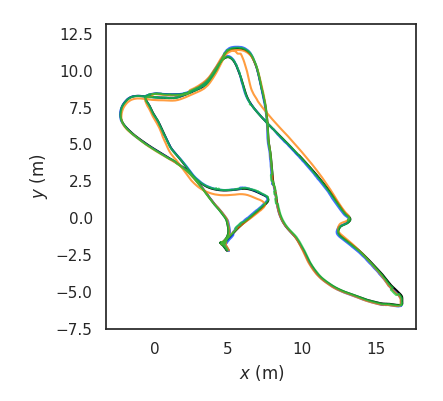} 
  \caption{MH05 Sequence}
\end{subfigure}
\begin{subfigure}{.19\textwidth}
  \centering
     \vspace{1cm}
  \includegraphics[width=0.98\linewidth, trim=1cm 0cm 0cm 1cm]{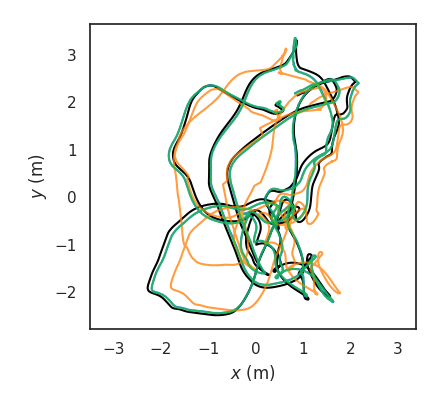} 
  \caption{V101 Sequence}
\end{subfigure}
\vspace{0cm}
\begin{subfigure}{.19\textwidth}
  \centering
     \vspace{1cm}
  \includegraphics[width=0.98\linewidth, trim=1cm 0cm 0cm 1cm]{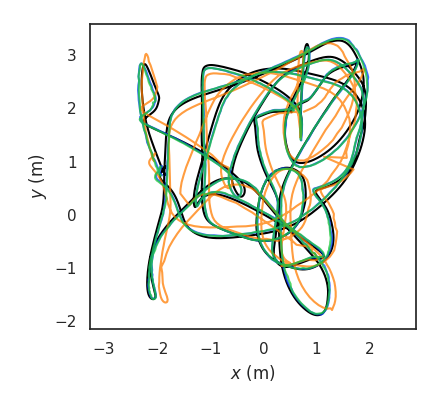} 
  \caption{V102 Sequence}
\end{subfigure}
\begin{subfigure}{.19\textwidth}
  \centering
     \vspace{1cm}
  \includegraphics[width=0.98\linewidth, trim=1cm 0cm 0cm 1cm]{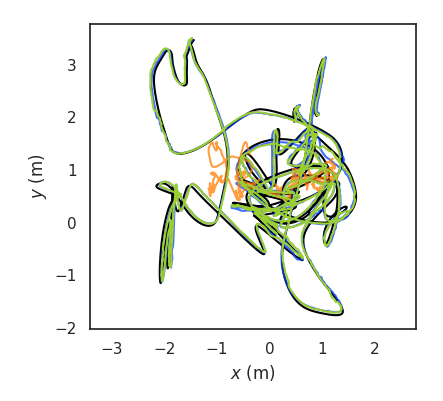} 
  \caption{V103 Sequence}
\end{subfigure}
\begin{subfigure}{.19\textwidth}
  \centering
     \vspace{1cm}
  \includegraphics[width=0.98\linewidth, trim=1cm 0cm 0cm 1cm]{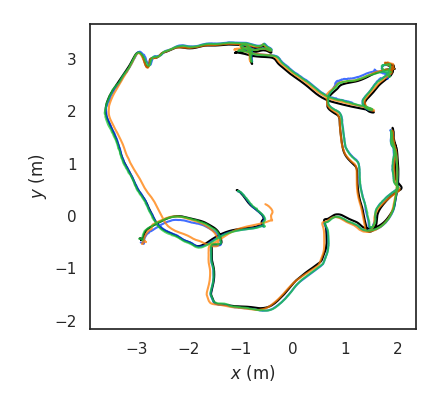} 
  \caption{V201 Sequence}
\end{subfigure}
\vspace{0cm}
\centering
\begin{subfigure}{.19\textwidth}
  \centering
     \vspace{1cm}
  \includegraphics[width=0.98\linewidth, trim=1cm 0cm 0cm 1cm]{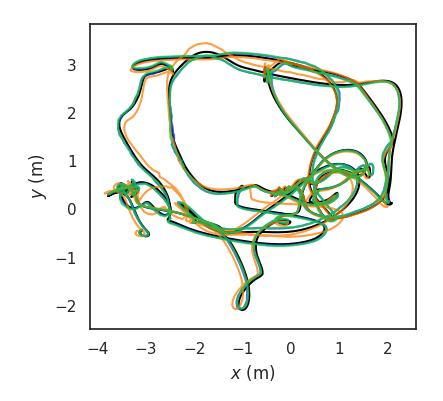} 
  \caption{V202 Sequence}
\end{subfigure}
\end{minipage}%
\vspace{0.3cm}
\caption{Trajectory in EuRoC dataset compared with ORB-SLAM2 and VINS-Stereo.}
\label{fig:traj_res}
\end{figure*}
\begin{figure*}[!htpb]
    \centering
\begin{minipage}{0.5\textwidth}
\centering
    \includegraphics[width=0.9\linewidth, trim=2.2cm 2.4cm 2cm 0cm]{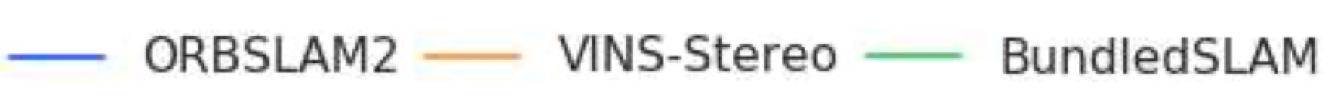}
\end{minipage}\\
\vspace{0.4cm}
\begin{minipage}{1\textwidth}
\begin{subfigure}{.19\textwidth}
  \centering
  \includegraphics[width=.98\linewidth, trim=1cm 0cm 0cm 1cm]{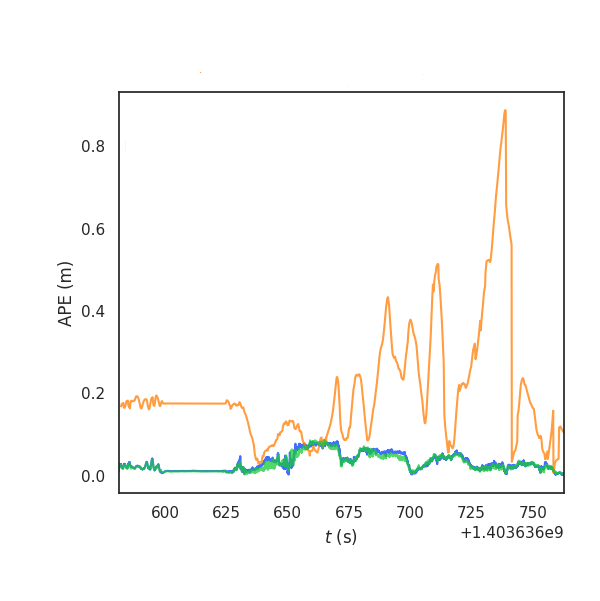} 
  \caption{MH01 Sequence}
\end{subfigure}
\begin{subfigure}{.19\textwidth}
  \centering
  \includegraphics[width=.98\linewidth, trim=1cm 0cm 0cm 1cm]{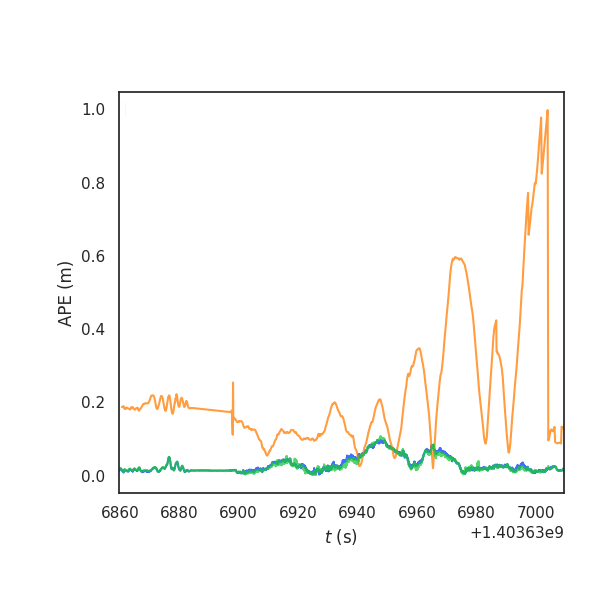} 
  \caption{MH02 Sequence}
\end{subfigure}
\begin{subfigure}{.19\textwidth}
  \centering
  \includegraphics[width=.98\linewidth, trim=1cm 0cm 0cm 1cm]{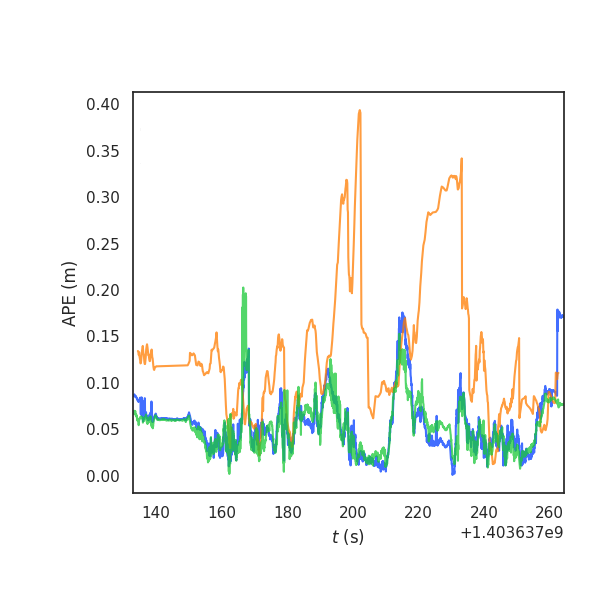} 
  \caption{MH03 Sequence}
\end{subfigure}
\begin{subfigure}{.19\textwidth}
  \centering
  \includegraphics[width=.98\linewidth, trim=1cm 0cm 0cm 1cm]{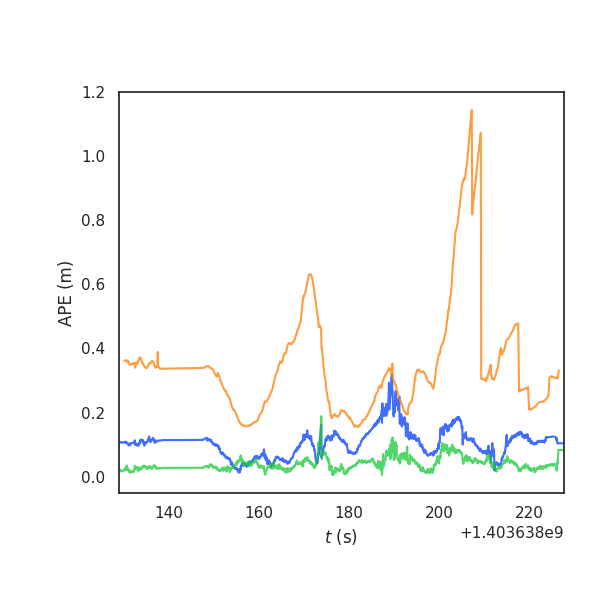} 
  \caption{MH04 Sequence}
\end{subfigure}
\begin{subfigure}{.19\textwidth}
  \centering
  \includegraphics[width=.98\linewidth, trim=1cm 0cm 0cm 1cm]{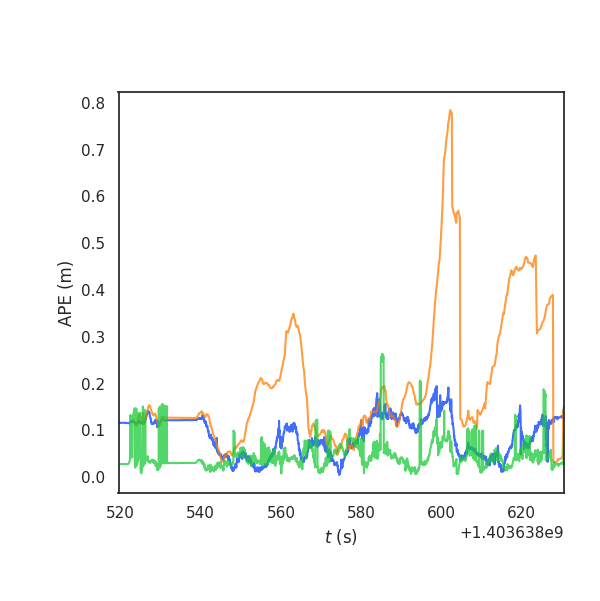} 
  \caption{MH05 Sequence}
\end{subfigure}
\end{minipage}%
\vspace{0.2cm}
\begin{minipage}{1\textwidth}
\begin{subfigure}{.19\textwidth}
  \centering
  \includegraphics[width=.98\linewidth, trim=1cm 0cm 0cm 1cm]{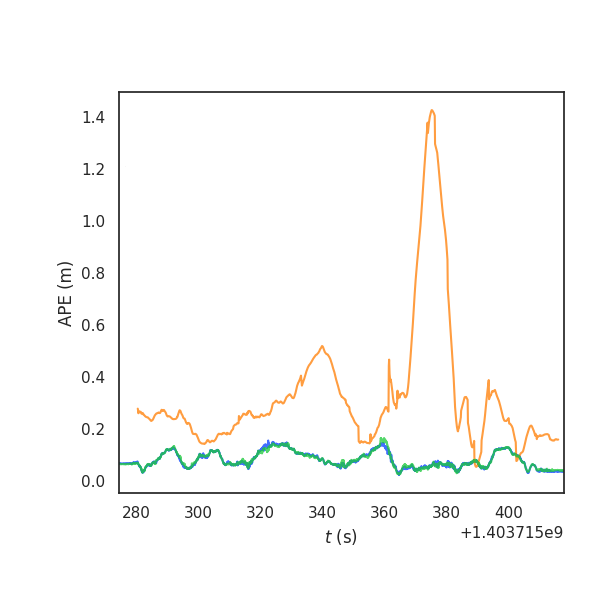} 
  \caption{V101 Sequence}
\end{subfigure}
\begin{subfigure}{.19\textwidth}
  \centering
  \includegraphics[width=.98\linewidth, trim=1cm 0cm 0cm 1cm]{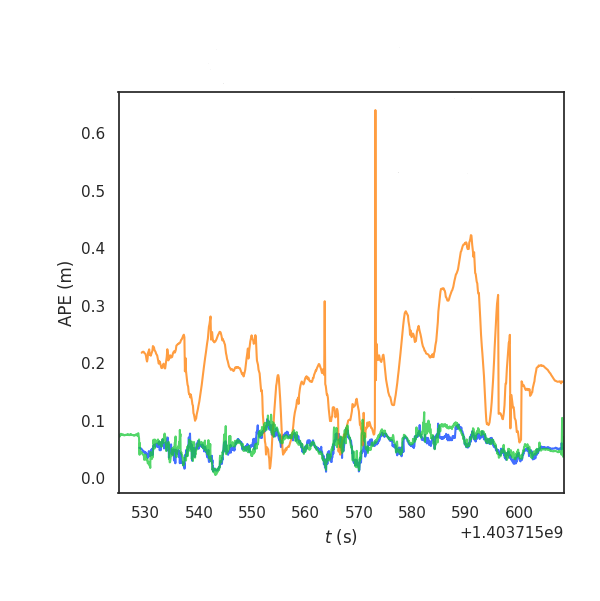} 
  \caption{V102 Sequence}
\end{subfigure}
\begin{subfigure}{.19\textwidth}
  \centering
  \includegraphics[width=.98\linewidth, trim=1cm 0cm 0cm 1cm]{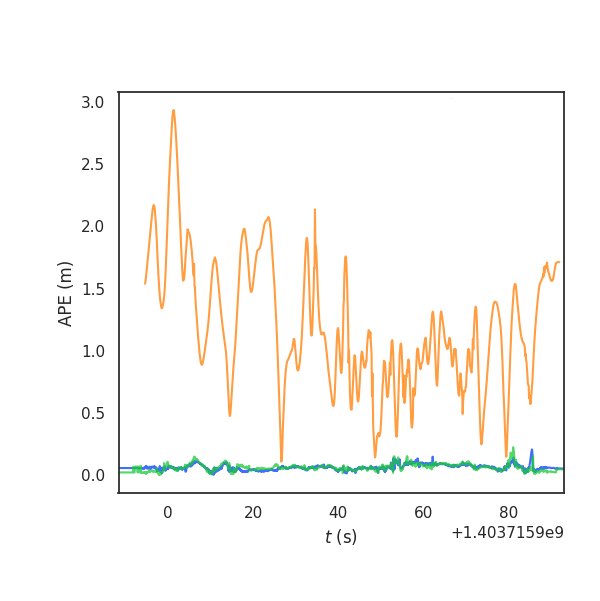} 
  \caption{V103 Sequence}
\end{subfigure}
\begin{subfigure}{.19\textwidth}
  \centering
  \includegraphics[width=.98\linewidth, trim=1cm 0cm 0cm 1cm]{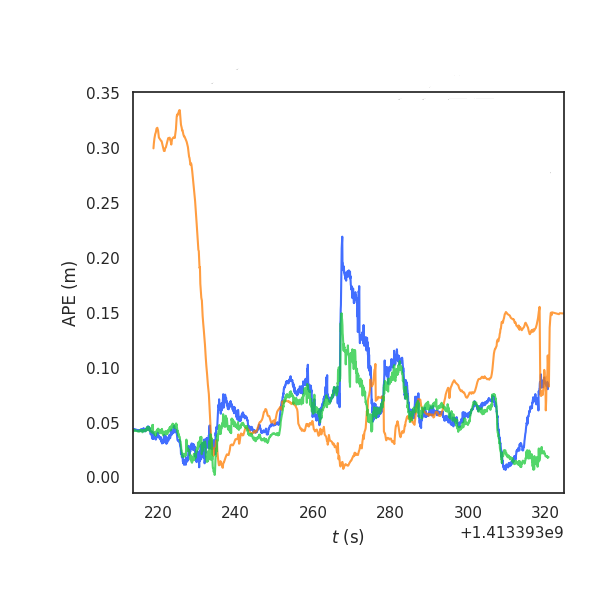} 
  \caption{V201 Sequence}
\end{subfigure}
\begin{subfigure}{.19\textwidth}
  \centering
  \includegraphics[width=.98\linewidth, trim=1cm 0cm 0cm 1cm]{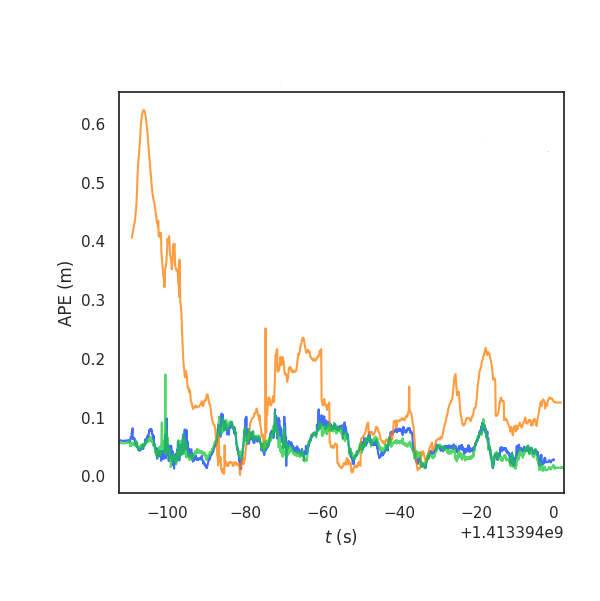} 
  \caption{V202 Sequence}
\end{subfigure}
\end{minipage}%
\vspace{0.2cm}
\caption{Comparison of Absolute Pose Error (APE) in EuRoC Dataset.}
\label{fig:ate_res}
\end{figure*}

\section{Evaluation}
\label{sec:eva}
Our BundledSLAM system operates on a PC equipped with an Intel Core i7-8700 (four cores @ 3.2GHz) processor and 16GB of RAM. 
To assess the performance of our system, we conducted evaluations using the EuRoc dataset~\cite{euroc}. We compared our system's performance to that of state-of-the-art SLAM systems, namely ORB-SLAM2 and VINS-Stereo, in order to highlight the accuracy of our system.
Considering the non-deterministic nature of multi-threading systems and the inherent randomness involved, we executed each sequence five times. This approach allows us to present not only the best results but also the median results, providing a comprehensive overview of the accuracy achieved in estimating the trajectory.
\subsection{EuRoC Datasets}
The EuRoC dataset was captured by a micro aerial vehicle (MAV) equipped with two global-shutter, monochrome cameras operating at 20 Hz, along with an inertial measurement unit (IMU) running at 200 Hz. This dataset comprises five sequences recorded in a spacious industrial machine hall and six sequences from two distinct rooms, each accompanied by precise ground truth data. These sequences are categorized into different sets based on factors such as lighting conditions, scene texture, and the speed of motion. These categories include easy, medium, and difficult sets.
To facilitate our multi-cameras system, we assume the extrinsic matrix between two cameras through calibration processes are known.
\subsection{Accuracy Evaluation}
\begin{table}[!tbp] 
\centering
{
\vspace{0.20in}
\caption{
Exhaustive initialization results for 10 keyframes with Low, Medium, and High Angular Velocity from V2\_03\_difficult sequence."
}
\label{tab:performance_comparison}
 \resizebox{\columnwidth}{!}{
\begin{tabular}{ccccccc}
    \toprule
    \multicolumn{1}{c}{\centering Sequence Id} & \multicolumn{2}{c}{\centering BundledSLAM} & \multicolumn{2}{c}{\centering ORB-SLAM2} & \multicolumn{2}{c}{\centering VINS-Stereo}\cr
    &Best&Average&Best&Average&Best&Average\cr
    \midrule
    MH\_01\_easy&{\bf 0.034}&{\bf 0.036}&0.038&0.039&0.228&0.235\cr
    MH\_02\_easy&{\bf 0.035}&{\bf 0.044}&0.042&0.046&0.267&0.276\cr
    MH\_03\_medium&{\bf 0.056}&{\bf 0.064}&0.058&0.066&0.148&0.149\cr
    MH\_04\_difficult&{\bf 0.046}&{\bf 0.092}&0.067&0.115&0.396&0.409\cr
    MH\_05\_difficult&0.057&{\bf 0.081} &{\bf 0.051}&0.089&0.249&0.255\cr
    V1\_01\_easy&{\bf 0.086}&{\bf 0.087}&{0.087}&{\bf 0.087}&0.405&0.412\cr
    V1\_02\_medium&{\bf 0.064}&{\bf 0.065}&{\bf 0.064}&{\bf 0.065}&0.201&0.205\cr
    V1\_03\_difficult&{\bf 0.068}&0.097&0.072&{\bf 0.090}&1.259&1.429\cr
    V2\_01\_easy&0.06&{\bf 0.061}&{\bf 0.056}&0.064&0.117&0.150\cr
    V2\_02\_medium&{\bf 0.054}&{\bf 0.056}&0.056&0.069&0.193&0.194\cr
    \bottomrule
    \end{tabular}
}
}
\end{table}
Table~\ref{tab:performance_comparison} provides an overview of the absolute translation errors (Root Mean Square Error - RMSE) for BundledSLAM across all sequences, as compared to ORB-SLAM2 and VINS-Stereo. These errors were computed after aligning the estimated trajectories with ground truth. Notably, we activated the loop closing module with global Bundle Adjustment (BA) for both our system and ORB-SLAM2 when processing sequences MH\_05\_difficult, V1\_02\_medium, V1\_03\_difficult, V2\_01\_easy, and V2\_02\_medium. However, due to the high motion speed in certain parts of sequence V2\_03\_difficult, both methods experienced difficulties. This challenge could potentially be mitigated by incorporating additional sensors, such as an Inertial Measurement Unit (IMU).
As a multi-camera system, BundledSLAM has demonstrated superior accuracy compared to the state-of-the-art system, ORB-SLAM2. It's worth noting that BundledSLAM and ORB-SLAM2 consistently outperform VINS-Stereo across all sequences. The top-performing method for each sequence is highlighted in \textbf{bold}.\par
Figure~\ref{fig:traj_res} provides a comparison between ORB-SLAM2, VINS-Stereo, and ground truth. It's evident that our multi-camera system consistently delivers more accurate estimates when tested on the EuRoC dataset. 
Furthermore, we conducted a comparison of the Absolute Pose Error (APE) between ORB-SLAM2, VINS-Stereo, and our BundledSLAM system, with the results depicted in Figure~\ref{fig:ate_res}. These results clearly demonstrate that our proposed system, BundledSLAM, consistently outperforms the others, consistently exhibiting the smallest APE for each sequence.

\section{Conclusion}
\label{sec:conclu}
This paper presents BundledSLAM, a visual SLAM system designed to harness the capabilities of multiple cameras. The system integrates data from various cameras into a unified "bundled frame" structure, facilitating real-time pose tracking, local mapping for pose and map point optimization, and loop closing to ensure global consistency. Our evaluation, conducted using the EuRoC dataset, highlights that our system consistently outperforms the original system, demonstrating its exceptional accuracy in both the best and average results.

In order to enhance system robustness, particularly in scenarios characterized by motion blur or limited texture features, we intend to explore sensor fusion, potentially incorporating Inertial Measurement Units (IMUs). However, we are aware of the computational complexity that additional sensors may introduce. As part of our future research, we will prioritize strategies to reduce this complexity while maintaining or even improving system performance.

\newpage
\bibliographystyle{IEEEtran}
\bibliography{references}

\begin{thebibliography}{10}
\providecommand{\url}[1]{#1}
\csname url@samestyle\endcsname
\providecommand{\newblock}{\relax}
\providecommand{\bibinfo}[2]{#2}
\providecommand{\BIBentrySTDinterwordspacing}{\spaceskip=0pt\relax}
\providecommand{\BIBentryALTinterwordstretchfactor}{4}
\providecommand{\BIBentryALTinterwordspacing}{\spaceskip=\fontdimen2\font plus
\BIBentryALTinterwordstretchfactor\fontdimen3\font minus \fontdimen4\font\relax}
\providecommand{\BIBforeignlanguage}[2]{{%
\expandafter\ifx\csname l@#1\endcsname\relax
\typeout{** WARNING: IEEEtran.bst: No hyphenation pattern has been}%
\typeout{** loaded for the language `#1'. Using the pattern for}%
\typeout{** the default language instead.}%
\else
\language=\csname l@#1\endcsname
\fi
#2}}
\providecommand{\BIBdecl}{\relax}
\BIBdecl

\bibitem{1211520}
R.~Pless, ``Using many cameras as one,'' in \emph{2003 IEEE Computer Society Conference on Computer Vision and Pattern Recognition, 2003. Proceedings.}, vol.~2, 2003, pp. II--587.

\bibitem{frahm2004pose}
J.-M. Frahm, K.~K{\"o}ser, and R.~Koch, ``Pose estimation for multi-camera systems,'' in \emph{Joint Pattern Recognition Symposium}.\hskip 1em plus 0.5em minus 0.4em\relax Springer, 2004, pp. 286--293.

\bibitem{4631502}
J.~Sola, A.~Monin, M.~Devy, and T.~Vidal-Calleja, ``Fusing monocular information in multicamera slam,'' \emph{IEEE Transactions on Robotics}, vol.~24, no.~5, pp. 958--968, 2008.

\bibitem{7}
A.~Harmat, I.~Sharf, and M.~Trentini, ``Parallel tracking and mapping with multiple cameras on an unmanned aerial vehicle,'' in \emph{Intelligent Robotics and Applications}, C.-Y. Su, S.~Rakheja, and H.~Liu, Eds.\hskip 1em plus 0.5em minus 0.4em\relax Berlin, Heidelberg: Springer Berlin Heidelberg, 2012, pp. 421--432.

\bibitem{8}
A.~Harmat, M.~Trentini, and I.~Sharf, ``Multi-camera tracking and mapping for unmanned aerial vehicles in unstructured environments,'' \emph{Journal of Intelligent {\&} Robotic Systems}, vol.~78, no.~2, pp. 291--317, May 2015.

\bibitem{9}
M.~J. Tribou, A.~Harmat, D.~W. Wang, I.~Sharf, and S.~L. Waslander, ``Multi-camera parallel tracking and mapping with non-overlapping fields of view,'' \emph{The International Journal of Robotics Research}, vol.~34, no.~12, pp. 1480--1500, 2015.

\bibitem{10}
S.~Yang, S.~A. Scherer, X.~Yi, and A.~Zell, ``Multi-camera visual slam for autonomous navigation of micro aerial vehicles,'' \emph{Robotics and Autonomous Systems}, vol.~93, pp. 116 -- 134, 2017.

\bibitem{trvomcs}
P.~Liu, M.~Geppert, L.~Heng, T.~Sattler, A.~Geiger, and M.~Pollefeys, ``Towards {R}obust {V}isual {O}dometry with a {M}ulti-{C}amera {S}ystem,'' \emph{IEEE/RSJ International Conference on Intelligent Robots and System}, 2018.

\bibitem{svo}
C.~Forster, Z.~Zhang, M.~Gassner, M.~Werlberger, and D.~Scaramuzza, ``{SVO}: {S}emi-{D}irect {V}isual {O}dometry for {M}onocular and {M}ulti-{C}amera {S}ystems,'' \emph{IEEE Transactions on Robotics}, vol.~33, no.~2, pp. 249--265, 2017.

\bibitem{ORBSLAM2}
R.~Mur-Artal and J.~D. Tardos, ``{ORBSLAM2}: an {O}pen-{S}ource {SLAM} {S}ystem for {M}onocular, {S}tereo and {RGB-D} {C}ameras,'' \emph{IEEE Transactions on Robotics}, vol.~33, no.~5, p. 1255–1262, 2017.

\bibitem{EDI}
W.~Wang, J.~Li, Y.~Ming, and P.~Mordohai, ``{EDI}: Eskf-based disjoint initialization for visual-inertial slam systems,'' in \emph{Proceedings of {IEEE/RSJ} International Conference on Intelligent Robots and Systems ({IROS})}, 2023.

\bibitem{covisib}
H.~Strasdat, A.~J. Davison, J.~M.~M. Montiel, and K.~Konolige, ``Double window optimisation for constant time visual {SLAM},'' \emph{IEEE International Conference on Computer Vision}, pp. 2352--2359, 2011.

\bibitem{GN}
B.~Triggs, P.~McLauchlan, R.~Hartley, and Fitzgibbon, ``Bundled adjustment - a modern synthesis,'' \emph{in Vision Algorithms: Therory and Pracitce. Springer Verlag}, pp. 298--375, 2000.

\bibitem{g2o}
R.Kuemmerle, G.~Grisetti, H.~Strasdat, K.~Konolige, and W.~Burgard, ``g2o: A general framework for graph optimization,'' \emph{International Conference on Robotics and Automation}, pp. 3607--3613, 2011.

\bibitem{DBoW}
D.~Galvez-Lpez and J.~D. Tardos, ``Bags of binary words for fast place recognition in image sequences,'' \emph{IEEE Transactions on Robotics}, vol.~28, no.~5, pp. 1188--1197, 2012.

\bibitem{vocab}
R.~Mur-Artal and J.~D. Tardos, ``Fast relocalisation and loop closing in keyframe-based slam,'' \emph{IEEE Internatinal Conference on Robotics and Automation}, pp. 846--853, 2014.

\bibitem{euroc}
M.~Burri, J.~Nikolic, P.~Gohl, T.~Schneider, J.~Rehder, M.~W.~A. S.~Omari, and R.~Siegwart, ``The {E}u{R}o{C} micro aerial vehicle datasets,'' \emph{The International Journal of Robotics Research}, vol.~35, no.~10, p. 1157–1163, 2016.

\end{thebibliography}

\end{document}